\documentclass[letterpaper, 10 pt, conference]{ieeeconf}  

\IEEEoverridecommandlockouts                              

\usepackage{amsmath,amssymb,amsfonts}
\usepackage{algorithmic}
\usepackage{graphicx}
\usepackage{textcomp}
\usepackage{xcolor}
\usepackage{apacite}

\usepackage{epstopdf}
\DeclareGraphicsExtensions{%
    .png,.PNG,%
    .pdf,.PDF,%
    .jpg,.mps,.jpeg,.jbig2,.jb2,.JPG,.JPEG,.JBIG2,.JB2}

\usepackage{subcaption}
\usepackage[ruled,vlined]{algorithm2e}
\SetKw{KwBy}{by}

\PassOptionsToPackage{hyphens}{url}\usepackage{hyperref}
\usepackage{multirow}
\usepackage[normalem]{ulem}

\makeatletter
\let\NAT@parse\undefined
\makeatother
\usepackage[comma]{natbib}  %
\renewcommand{\BBA}{} 
\setcitestyle{authoryear,open={(},close={)},aysep={}}

\title{\LARGE \bf
Analyzing Neural Jacobian Methods in Applications of Visual Servoing and Kinematic Control
}

\author{Michael Przystupa$^{\dagger}$, Masood Dehghan$^{\dagger}$, Martin Jagersand$^{\dagger}$ and A. Rupam Mahmood$^{\dagger *}$
\thanks{$^{\dagger}$Department of Computing Science,
        University of Alberta, Edmonton AB., Canada, T6G 2E8.
        { 
           \tt\small \{przystup, masood1, mj7, armahmood\}@ualberta.ca
        }
        *CIFAR AI Chair, Alberta Machine Intelligence Institute (Amii)
        }%
}

\begin{document}

\maketitle
\thispagestyle{empty}
\pagestyle{empty}

\begin{abstract}
Designing adaptable control laws that can transfer between different robots is a challenge because of kinematic and dynamic differences, as well as in scenarios where external sensors are used. In this work, we empirically investigate a neural networks ability to approximate the Jacobian matrix for an application in Cartesian control schemes. Specifically, we are interested in approximating the kinematic Jacobian, which arises from kinematic equations mapping a manipulator's joint angles to the end-effector's location. We propose two different approaches to learn the kinematic Jacobian. The first method arises from visual servoing where we learn the kinematic Jacobian as an approximate linear system of equations from the k-nearest neighbors for a desired joint configuration. The second, motivated by forward models in machine learning, learns the kinematic behavior directly and calculates the Jacobian by differentiating the learned neural kinematics model. Simulation experimental results show that both methods achieve better performance than alternative data-driven methods for control, provide closer approximations to the proper kinematics Jacobian matrix, and on average produce better-conditioned Jacobian matrices. Real-world experiments were conducted on a Kinova Gen-3 lightweight robotic manipulator, which includes an uncalibrated visual servoing experiment, a practical application of our methods, as well as a 7-DOF point-to-point task highlighting that our methods are applicable on real robotic manipulators.

\end{abstract}

\section{Introduction} \label{sec:introduction}

This work focuses on learning the kinematics Jacobian, which arises from a manipulator's forward kinematic equations. Forward kinematics describes the location of a manipulator's end-effector $x = f(q)$ where $f$ is the manipulator's kinematic equations, $x\in \mathbf{R}^m$ is the end-effector, and $q\in\mathbf{R}^n$ are the joint angles. Differentiating with respect to time describes the velocity relation between joints and end-effector $J(q)\dot{q} = \dot{x}$ where $J(q) \in R^{m \times n}$ is the kinematic Jacobian. In robotic applications, the kinematic Jacobian can be useful for controlling robots, such as in Cartesian control schemes \citep{craig1989IntroRobotics}. These classical controllers are often preferred in practice because of their well-understood behavior. Fundamentally, Cartesian control seeks to solve an ill-posed inverse problem and other works have explored directly learning input gradients \citep{adler2017solvillposedinv,song2020human}, but ignore analyzing the intermediate Jacobian representation 

However, calculating the true Jacobian can be difficult in practice for less controlled environments. One reason is due to uncertainty in the kinematics or dynamics of the robotic system. Uncertainty can arise in object manipulations with robotic fingers where all fingers impose complex-to-model parallel-kinematics \citep{jagersand1996acquiringvismodel}. Furthermore, the introduction of external sensory information changes the structure of the Jacobian. An example of this is visual servoing (VS), where an analytic Jacobian exists but can be difficult to calculate in practice \citep{hitchinson1996tutorialVS}. In these situations, an adaptive learning approach is beneficial and can help enable the application of robots in less controlled settings \citep{hitchinson1996tutorialVS,jun2019visualgeometric,Jun_ICRA2019}. 

Jacobian approximation methods, both local and global approaches, have been previously studied in uncalibrated visual servoing (UVS) research. Local methods typically use a finite difference approximation that can be updated online \citep{jagersand96visualservoing,jagersand1997experimentalevalUVS,piepmeir2004UDVS,qian2002kalmanuvs}. These methods are convenient as they are easy to interpret and have local convergence guarantees \citep{nematollahi2009generalizedcontrolawsIBVS}. However, a limitation is they do not remember the information collected while following a trajectory and fail to converge toward distant targets.  Global methods mitigate this by incorporating information from the entire workspace to increase performance stability. However, to the best of our knowledge, current research has only considered non-parametric methods, such as K-nearest neighbors, which can be memory intensive \citep{farahmand2007globaluvs}. We propose that neural networks can be effective Jacobian approximators that work well compared to alternative UVS techniques. Specifically, we propose two methods based on neural networks trained to predict the Jacobian matrix, specifically with an application for kinematic control schemes in mind.


Our first method predicts the Jacobian matrix directly by utilizing a loss function based on the equations of motion for robots. It is motivated by previous work in uncalibrated visual servoing \citep{farahmand2007globaluvs}. We refer to this as the \textit{Neural Jacobian} method. To our knowledge, our Neural Jacobian formulation is novel. However, variations have previously been proposed in the literature \citep{zhao2009NNmultifingerJacobian,lyu2018NNvisionjacobianpred,lyu2020datadrivenLearningofJacobian}. Our variation of the method only requires access to position information. In contrast to other works that focus more on theoretical understanding, we focus on the empirical evaluations of our approach.

The second method predicts the forward kinematics and the Jacobian is recovered with back-propagation. We refer to this forward model approach as the \textit{Neural Kinematics} method. This approach is motivated from deep learning theory, which suggests that neural networks can effectively approximate derivative information \citep{hornik1990universalapproximatorandderivs,hornik95degreeof,maiduy200rbfderivapprox}. Approximating gradients in sub-networks appears across various machine learning fields such as generative adversarial networks \citep{goodfellow2014GANS} and reinforcement learning \citep{lillicrap2016ddpg,gal2016improving}. Most closely tied to our Neural Kinematics approach is the work of Jordan et al. (\citeyear{jordan1989GenericConstraints}) where Jacobian approximation was explicitly discussed. Our work differs as we directly use the approximate Jacobian for control, whereas they use it to optimize a sub-network for control more typical in deep learning.  

In our experiments, both methods are studied in an \textit{ab initio learning} setting where the robot does not have a-priori knowledge of it's own rigid body and must learn to control itself through interacting with the environment \citep{Censi2013BootstrappingV}. We evaluate our controllers on set-point reaching tasks in extensive simulation experiments. The objective of these tasks is to move our robot's end-effector towards a target $x^{*} \in \mathbf{R}^{m}$. We analyze several aspects of the Jacobian matrices from our control tasks and find that our neural methods are generally closer approximations to alternative learning methods and can be better behaved on average than the alternative ab initio approaches considered. We then provide proof-of-concept applications of our methods on a Kinova Gen-3 light robotic manipulator in a 2-DOF visual servoing task, as well as a  kinematic control experiment using all 7-DOF of the manipulator \citep{kinova}. We focus on the general case of kinematic control and do not study singularity configurations.  To our knowledge, singularities have non-trivial problems warranting special consideration, and are left for future works \citep{consistentNullspac1993Chen,overviewNullspace2015dietrich}.

In Section~\ref{sec:background} we provide pertinent background information. Section~\ref{sec:methodology} discusses our proposed methods in detail. In Section~\ref{sec:experiments} we report our experimental results in simulation and on the Kinova manipulator. We conclude and discuss future research directions in Section~\ref{sec:conclusion}.

\begingroup
\begin{figure*}[h]
\vspace{0.1in}
\hspace{0.07in}
    \begin{minipage}[t]{0.33\textwidth}
        \centering
        
        \includegraphics[trim=10mm 15mm 10mm 10mm, clip=true, width=\textwidth]{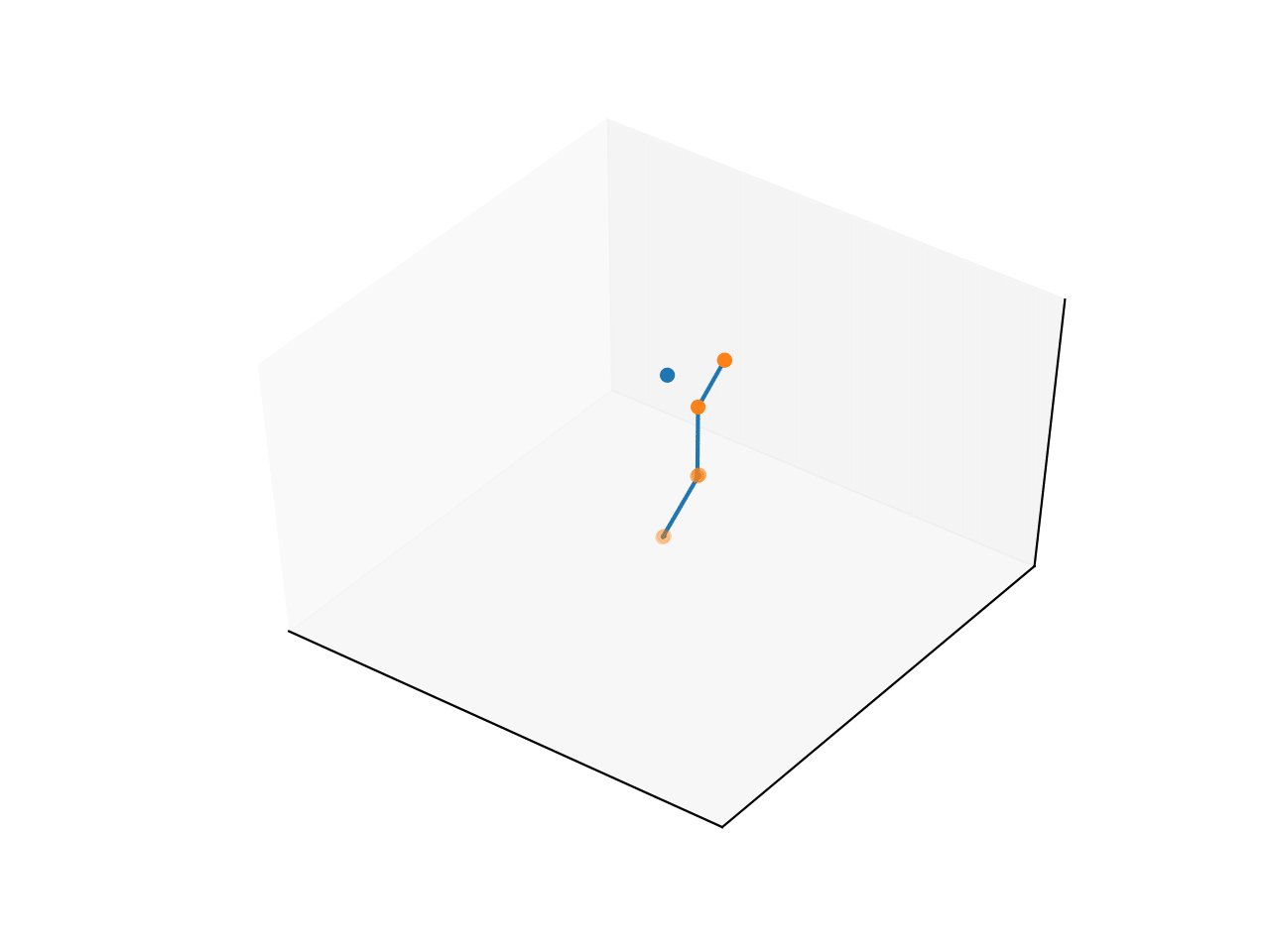}
        \caption*{Single Point Simulator.}
        \label{fig:1ptsimulator}
    \end{minipage}
    \begin{minipage}[t]{0.33\textwidth}
        \centering
        
        \includegraphics[trim=10mm 15mm 10mm 10mm, clip=true, width=\textwidth]{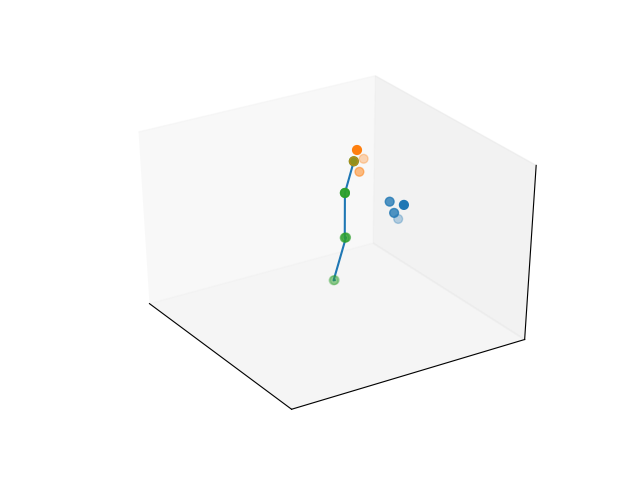} 
        \caption*{Multiple Point Simulator.}
        \label{fig:4ptsimulator}
    \end{minipage}
    \begin{minipage}[t]{0.33\textwidth}
    \centering
        \includegraphics[trim=10mm 25mm 10mm 10mm, clip=true, width=0.9\textwidth]{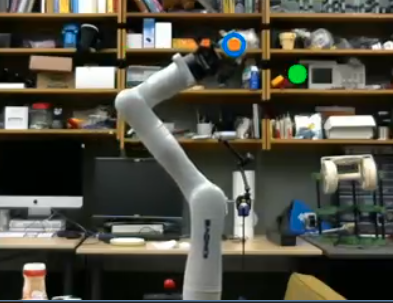}
        \caption*{Kinova Manipulator Robot. }
        \label{fig:kinova}
    \end{minipage}
    
\caption{Robotic Environments for Experiments. The multi-point environment add's three points which correspond to the orientation of the end-effector as defined in Cartesian space. For the visual servoing task on the Kinova, green is the target and the blue circle is the tracked position.}
\label{fig:experiments}
\end{figure*}
\endgroup

\section{Background} \label{sec:background}
In this section, we briefly discuss relevant topics for understanding our work. The topics we considered most relevant for comprehension are forward kinematics, Cartesian control laws, and neural networks. 

\subsection{Forward Kinematics}
A robot can be represented by a collection of rigid body links and joint angles between these links. A significant relationship between links and joint angles is the location of the tip of the final link, or end-effector, in relation to the base of a robot. 
This relationship can be modeled with forward kinematics, which describes the relationships between joint angles and links with homogeneous transformations ${}^{i-1}T_{i} \in \mathbf{R}^{4\times4}$, where $i-1$ refers to the joint we are transforming coordinate frames from to joint $i$. Chaining these transformations together then describes the change of coordinate systems from base frame to end-effector's coordinate system, and can be viewed as $f(q) = x$ where $q \in R^{n}$ is our joint angles and $x \in R^{m}$ is our end-effector location where the dimensions $m$ and $n$ may not be equal. In visual servoing, the end-effector location $x$ includes an additional projection function mapping it to the image plane surface \citep{hitchinson1996tutorialVS}.

\subsection{Cartesian Control Scheme}

The relationship between the motion of a robot's end-effector and joint angle velocities can be described with the following equation $\dot{x} = J(q) \dot{q}$, where $\dot{x} = \frac{\delta x}{\delta t}$ and $\dot{q} = \frac{\delta q}{\delta t}$ are the end-effector and actuator velocities, respectively. From this equation the \textit{inverse Jacobian} control law can be derived by multiplying through by $\delta t$ on both sides and replacing the resulting finite difference in position $\Delta x$ with $x^{*} - x$  to represent our target position. This substitution comes from formulating the control law as a problem solvable with Newton's method \citep{heath1996scientificcomputing}.

For each step of control, the control output joint velocities are then calculated as:
\begin{equation} \label{eqn:inversejacobian}
    \Delta{q} = \lambda [ J^{\dagger} (x^{*} - x) + (I - J^{\dagger}J)y ].
\end{equation}
where, $\lambda$ refers to the gain parameter for the corresponding velocity to be sent to the robot and $J^{\dagger} = (J J^{T})^{-1} J^{T}$ denotes the Moore-Penrose inverse of $J$. Equation~(\ref{eqn:inversejacobian}) is interpreted as sending a joint velocity command $\dot{q}$ \citep{craig1989IntroRobotics}. Algorithm~\ref{alg:inversejacobianpolicy} shows pseudo-code for this control law. The $(I - J^{\dagger}J)y$ term is included for completeness of the control law in the case of redundancy in a manipulator or tracked points in visual servoing \citep{consistentNullspac1993Chen,overviewNullspace2015dietrich}, where $y$ is a vector solution of the null space. In our experiments, this second term is not included as part of the evaluation. 

\begin{algorithm}[]
\SetAlgoLined

 Inputs: joint angles $q$, Target position $x^{*}$, current position $x$\;
 Result: velocity command $\dot{q}$\;
$\Delta x \leftarrow x^{*} - x$  \qquad \qquad  distance to target\; 
 $J \leftarrow J_{\theta}(q)$ \qquad \qquad  evaluate Jacobian\;
 $\Delta q \leftarrow \lambda [ J^{\dagger} \Delta x + (I - J^{\dagger}J)y$ \qquad  robot command\;
 
  \caption{Inverse Jacobian Controller }
  \label{alg:inversejacobianpolicy}
\end{algorithm}

\subsection{Neural Networks}

Neural networks are a type of function approximator constructed by a series of stacked matrix products with intermittent non-linear functions $\sigma$ between layers. They are composed of $N$ layers of learned parameters where $\{W^{i}, b^{i}\}$ correspond to layer $i$'s learning parameters $W^{i} \in \mathbf{R}^{m^{i}\times n^{i}}$ with a bias term $b^{i} \in \mathbf{R}^{n^{i}}$. Here, $n^{i}$ refers to the dimension of the input to a specific layer, and $m^{i}$ is the dimension of the outputs of that same layer. They need not be equal for each layer. Throughout this paper when referring to a neural function we use $\theta$ to refer to the networks parameters. As an example, here is a simple two hidden neural network for regression: $f_{\theta}(q) = W^{3}\sigma(W^{2}\sigma(W^{1}q + b^{1}) + b^{2}) + b^{3}$. These networks are typically optimized with back-propagation using a stochastic gradient descent algorithm \citep{goodfellow2016deeplearning}. 

\begin{figure}

    \centering
  \includegraphics[width=0.95\linewidth]{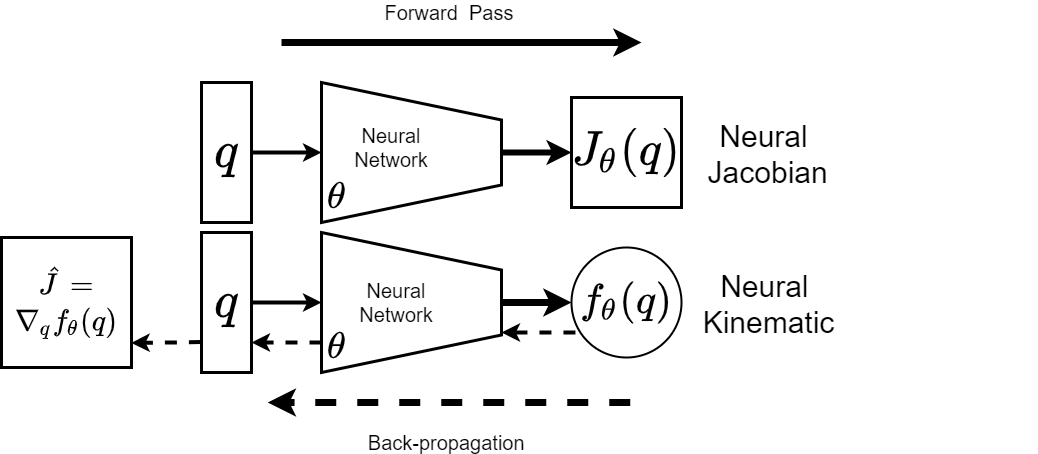}  
  
\caption{Deployment-time computations: The Neural Jacobian directly predicts the Jacobian, where as the Neural Kinematics calculates the Jacobian with respect to predicted position.}

    \label{fig:neuralmethods}
\end{figure}
\section{Methodology} \label{sec:methodology}

In this section, we describe both of our proposed methods for Jacobian approximation. These include the Neural Jacobian and Neural Kinematics methods, which we visualize the deployment-time computation in Figure~\ref{fig:neuralmethods}. The Neural Jacobian directly predicts the Jacobian. The Neural Kinematics approach instead predicts the kinematics relationship, and the Jacobian is predicted by back-propagation. 

\subsection{Neural Jacobian}

Our Neural Jacobian method directly predicts the Jacobian and is trained with supervised learning. We assume there is only access to joint angle and end-effector locations. To define an objective, we make a linearization assumption by enforcing the Secant condition  $\Delta x_{ij} \approx J(q_{t}) \Delta q_{ij}$, where for points $i$ and $j$ $\Delta x_{ij} = x_{i} - x_{j}$ and similarly for $\Delta q_{ij}$. Each pair $(\Delta x_{ij}, \Delta q_{ij})$ is formed by the finite differences of the $k$-nearest neighbors of the current $q$  being predicted \citep{jagersand96visualservoing}. We then fit our Jacobian on this set of finite difference: 
\begin{equation} \label{eqn:hyperplanejacobian}
    \min_{\theta} \sum_{i=1}^{k}\sum_{j=1}^{k}|| \Delta x_{ij} - J_{\theta}(q) \Delta q_{ij} ||_{2}^{2} + \beta || \Delta q_{ij} -  J_{\theta}^{\dagger}(q) \Delta x_{ij}||_{2}^{2}.
\end{equation}
In this objective, the first term was originally proposed by the work of Farahmand et al.  (\citeyear{farahmand2007globaluvs}) where they fit a multi-output linear regression model and treat the weights as the Jacobian. We extend the objective in this work with the second term weighted by $\beta$ which models the inverse relation; we assume that $J^{\dagger}$ is differentiable \citep{golub1973differentiatingInverse}. The intuition is that by training the network to predict both the forward and inverse relation, our Jacobian would be better conditioned when inverted for control. Our experiments suggest that this bi-directional learning's utility depends on whether the systems of equations are over or under constrained. In our experiments, when $\beta=0.0$ we refer to this as the Neural Jacobian and $\beta=1.0$, we call it the Bi-directional Neural Jacobian. We pre-compute the $k$-nearest neighboring finite differences of each $(q, x)$ before training the models. 

\subsection{Neural Kinematics}

As previously discussed, we can typically assume some optimal function $f^{*}$ representing the true relationship between the robot's actuator values $q$ and end-effector $x$. If we knew $f^{*}$, we could recover the Jacobian by directly differentiating with respect to the model. With this observation, the Neural Kinematics method approximates the kinematics relationship $f_{\theta}(q) = \hat{x}$ as opposed to the Jacobian in order for us to differentiate with an approximate kinematics model. We can optimize the Neural Kinematics model with supervised learning with any appropriate loss on training examples $(x, q)$ from the robot system. In our experiments we use the mean squared error $|| x - f_{\theta}(q)||^{2}_{2}$. 

After training the model, we freeze the Neural Kinematic network's weights and use back-propagation to recover the Jacobian. We define $f^{i}_{\theta}(q)$ as predicting a particular dimension of the Cartesian coordinates of $x$. The Jacobian can then be calculated by differentiating with respect to $q$ for each output dimension, i.e. ${\hat J_{ij}} = \frac{\partial f_{\theta}^{i}}{\partial q_{j}}$. Intuitively, if a function is well approximated, differentiating with respect to the joint angles against the neural network predictions should approximate the Jacobian matrix; theoretical results support this intuition \citep{hornik1990universalapproximatorandderivs,hornik95degreeof}.

\section{Experiments} \label{sec:experiments}

For our experiments, we consider several set-point control tasks both in simulation and on a Kinova Gen3 Light robotic manipulator \citep{kinova}. As part of our simulation work, we analyzed the Jacobian approximations to empirically probe how good the approximations are and identify when they might fail. These analyses are helpful for neural networks because directly analyzing the network's weights can be challenging to interpret.

We follow a similar experimental set-up for all of our control experiments. We generated trajectories of 100 discrete time steps generated by random joint velocity commands produced by an Ornstein and Uhlenbeck process \citep{uhlenbeck1930brownianmotion}. The one exception is our evaluations on the 7-DOF Kinova experiment where we instead generated commands with the true Jacobian inverse Jacobian controller perturbed by small Gaussian noise $\dot{q} + \epsilon$ ($\epsilon \sim N(0, 0.1)$) for 5\% of commands. The robot and simulator are reset to the same initial position for each collected trajectory to bias the data collection. This decision was intentional as it meant we had a greater density of data in parts of the work-space and fewer points in other parts showing how well our approximations might generalize. We hold out 15\% of training data for a validation set that selects the best network for deployment. All our networks use hidden layers of 100 neurons and differ only in the number of hidden layers: 2 layers for 7-DOF single point tasks (simulation and Kinova), one layer for 2-DOF visual servoing, and four layers for the multiple point simulation experiment. We also varied the amount of collected training data: 100,000 examples for 7-DOF single point tasks (simulation and Kinova), 10,000 for visual servoing task, and 200,000 for the multiple point simulation. We optimize our neural networks with the Adam optimizer with learning rate of 0.001, $\beta_{1}=0.9$, $\beta_{2}=0.999$ \citep{kingma2014method}.  

In our results, we compare our neural models against several baselines, including the True Kinematic Jacobian (TJ), Broyden's method \citep{jagersand96visualservoing}, and the Local Linear $k$-nearest-neighbor (LL-KNN) approach with $k=128$ in simulation and $k=50$ on the Kinova \citep{farahmand2007globaluvs}. We emphasize that comparing to the true Jacobian is for perspective on desirable performance for data-driven approaches in Cartesian control. We use a neighborhood $k = 10$ for our Neural Jacobian (NJ) and Bidirectional (Bi-NJ) algorithms, and ReLU activation functions for both. We did find some performance differences when varying the neighborhood size, but we report results on $k=10$, which was our initial choice and did not perform much worse than other choices. We report our Neural Kinematics results with both Tanh (Tanh-NK) and ReLU (Relu-NK) activation functions because of the notable differences in performance. For both NK variants in simulation, we considered differing amounts of weight decay during training. In the single-point tasks (simulated and Kinova), these values were 0.0 for Tanh-NK and 1e-04 for Relu-NK. In the multi-point simulator, these were 1e-06 for Tanh-NK and 1e-05 for Relu-NK. We note that in the weight decay ranges we evaluated, performance gains were larger with Relu-NK.\footnote{We make our code available here: \url{https://github.com/gamerDecathlete/neural_jacobian_estimation}}

\subsection{Simulation Control Experiments}
We first evaluate our neural algorithms in kinematic simulations of the Kinova where the position dynamics are determined as $q_{t+1} = q_{t} + \Delta t \dot{q}$. We consider a point-to-point (single point) and multiple point configuration, where the three additional points are a scaled coordinate frame shifted to the end-effector location (multi-point) task both featured in Figure~\ref{fig:experiments}. The 4 points of the multi-point environment represent the orientation and position of the end-effector in Cartesian space. In both settings, we collected 110 targets generated randomly from 10 different random seeds for 1100 trajectories in total to evaluate our methods. 

\begin{figure}[ht]
\hspace{.05in}
  \includegraphics[width=.95\linewidth]{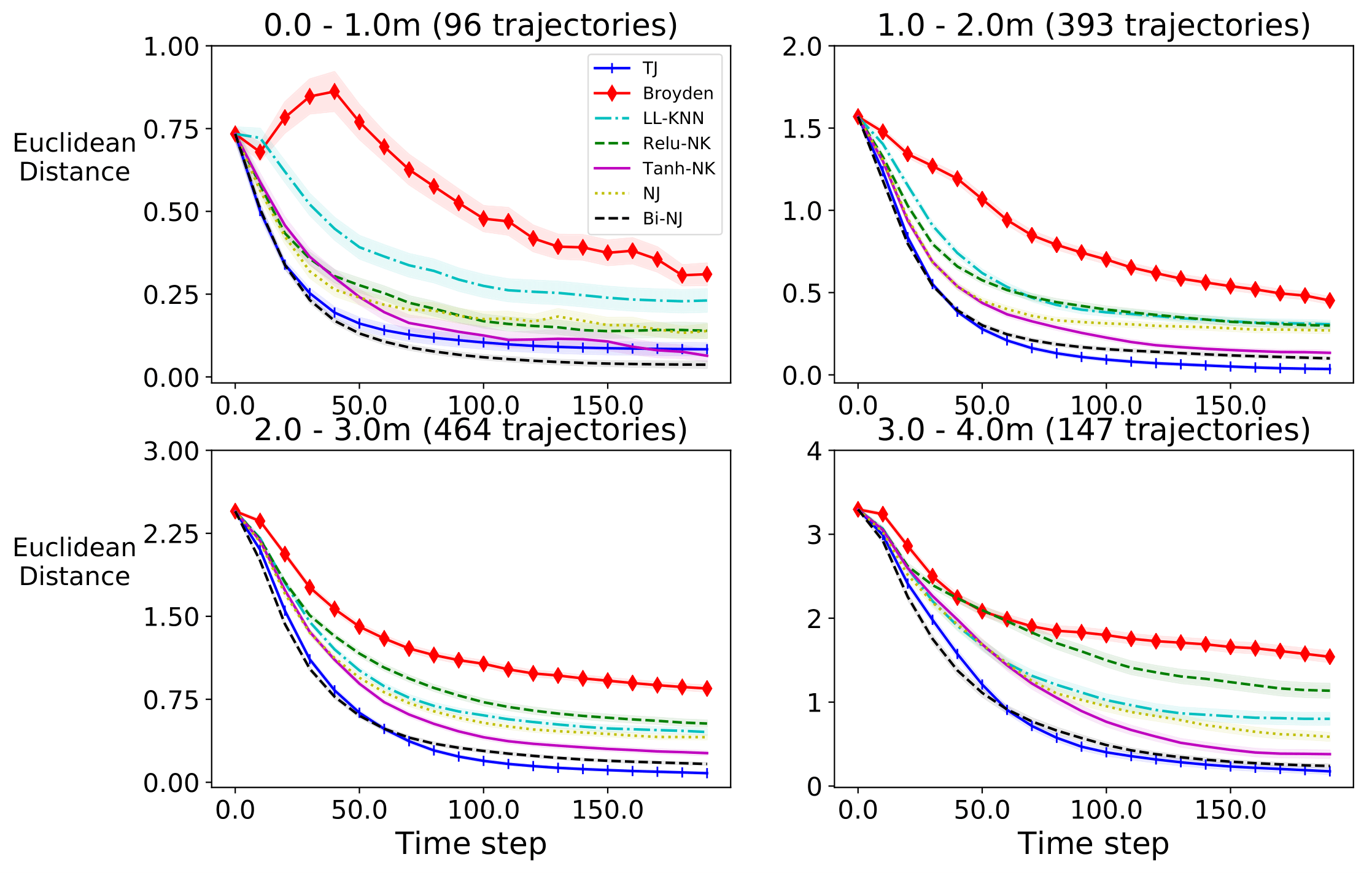}  
\caption{Euclidean distance (meters) to target with different control algorithms in simulation for multi-point environment. Plotted are the average performances across trajectories separate by initial starting distances with standard-error-of-mean. The distance is the sum of each point's distance to their respective targets.}
\label{fig:euclideandist}
\end{figure}

Figure~\ref{fig:euclideandist} shows the decrease in Euclidean distance from the initial position to the target. These plots are the average performance for trajectories from a range of starting distance with the standard-error over the number of trajectories. Due to space constraints, we only report the multi-point results where performance differences were more pronounced. In the multi-point setting, we found more divergence in behavior, particularly for more distant points. We accredit this to our biased data collection; there were less data around these distal points, so our approximations are less accurate.

\begin{table}[ht]
\centering
\setlength{\tabcolsep}{3 pt}

\begin{tabular}{c|c|c|c|c|c|}
\cline{2-6}
                                       & \multicolumn{5}{c|}{\textbf{Single Point Initial Distance (meters)}}                                   \\ \hline
\multicolumn{1}{|c|}{\textbf{Model}}   & \textbf{0.0 - 0.5} & \textbf{0.5 - 1.0} & \textbf{1.0 - 1.5} & \textbf{1.5 - 2.0} & \textbf{0.0 - 2.0} \\ \hline
\multicolumn{1}{|c|}{\textbf{Broyden}} & 99.91              & 61.52              & 35.49              & 3.37               & 47.84              \\ \hline
\multicolumn{1}{|c|}{\textbf{LL-KNN}}  & 92.29              & 87.11              & 79.97              & 60.00              & 81.58              \\ \hline
\multicolumn{1}{|c|}{\textbf{NJ}}      & \textbf{100.00}    & 92.96              & 82.89              & 59.77              & 85.70              \\ \hline
\multicolumn{1}{|c|}{\textbf{Bi-NJ}}   & 99.74              & 87.92              & 73.29              & 48.86              & 78.57              \\ \hline
\multicolumn{1}{|c|}{\textbf{Relu-NK}} & 99.13              & 89.83              & 84.91              & 72.84              & 86.76              \\ \hline
\multicolumn{1}{|c|}{\textbf{Tanh-NK}} & \textbf{100.00}    & \textbf{96.28}     & \textbf{91.94}     & \textbf{89.49}     & \textbf{94.07}     \\ \hline
\multicolumn{1}{|c|}{\textbf{TJ}}      & 100.00             & 97.89              & 94.27              & 90.93              & 95.81              \\ \hline
\end{tabular}
\caption{Percentage of successfully reaching targets for different methods. Results are arithmetic mean of success percentage for thresholds from 0.001m - 0.1m from target for single-point increment step-size of 0.001. In bold is the best performing method for a given distance.}
\label{tab:successtofailpointtopoint}
\end{table}

\begin{table}[ht]
\centering
\setlength{\tabcolsep}{3 pt}

\begin{tabular}{c|c|c|c|c|c|}
\cline{2-6}
                                       & \multicolumn{5}{c|}{\textbf{Multiple Points Initial Distance (meters)}}                                \\ \hline
\multicolumn{1}{|c|}{\textbf{Model}}   & \textbf{0.0 - 1.0} & \textbf{1.0 - 2.0} & \textbf{2.0 - 3.0} & \textbf{3.0 - 4.0} & \textbf{0.0 - 4.0} \\ \hline
\multicolumn{1}{|c|}{\textbf{Broyden}} & 38.45              & 24.59              & 10.50              & 1.60               & 16.78              \\ \hline
\multicolumn{1}{|c|}{\textbf{LL-KNN}}  & 59.78              & 54.93              & 45.82              & 30.01              & 48.18              \\ \hline
\multicolumn{1}{|c|}{\textbf{NJ}}      & \textbf{63.88}     & 50.04              & 35.56              & 24.00              & 41.66              \\ \hline
\multicolumn{1}{|c|}{\textbf{Bi-NJ}}   & \textbf{89.54}     & 72.45              & 59.13              & 38.55              & 63.79              \\ \hline
\multicolumn{1}{|c|}{\textbf{Relu-NK}} & 68.10              & 52.71              & 42.80              & 20.81              & 45.61              \\ \hline
\multicolumn{1}{|c|}{\textbf{Tanh-NK}} & \textbf{85.82}     & \textbf{76.44}     & \textbf{64.08}     & \textbf{49.21}     & \textbf{68.41}     \\ \hline
\multicolumn{1}{|c|}{\textbf{TJ}}      & 81.78              & 91.94              & 82.40              & 78.79              & 85.27              \\ \hline
\end{tabular}
\caption{Percentage of successfully reaching targets for different methods. Results are arithmetic mean of success percentage for thresholds from 0.001m - 0.25m for multi-point with an increment step-size of 0.001. Best performances are bold. Distance is the sum of distances between 4 points to their corresponding targets.}
\label{tab:successtofailmultiplepoint}
\end{table}

To distinguish algorithm performances, we separated trajectories into success or failure by a threshold on the final Euclidean distance to the target. We averaged performance over thresholds from 0.001m - 0.1m for the single-point experiments. For the multi-point experiments, we averaged from 0.001 - 0.25, which is more prominent because the distance is accumulative of all points. We used an increment size of 0.001m for each experiment.  We averaged performance to account for varying performance results observed when choosing a single threshold value. Table~\ref{tab:successtofailpointtopoint} shows results for single point reaching and Table~\ref{tab:successtofailmultiplepoint} shows results for multiple points experiments. In the single point case, all neural methods performed better than our learning baselines. In the multi-point case, we found hyperparameter choices had a greater effect on performance. The use of tanh activation worked better for most distances for the NK method. We gained notable improvement with the Bi-directional Neural Jacobian, even exceeding the true Jacobian results for closer points in the Neural Jacobian case. We conjecture that the Bi-directional neural Jacobian method's performance gain in the multi-point case is because fitting the system of equations is over-determined as opposed to under-determined in the single point case.   

\subsection{Jacobian Analysis}
To probe the performance results in our simulated control experiments, we analyzed several aspects of the Jacobian, including the Frobenius differences between approximate $\hat{J}$ and the true Jacobian $J^{*}$, the matrix condition number, and empirically checking the Lypanov convergence criteria. These metrics offer some insight into our Jacobian approximations, we recognize there are additional ways we could have analyzed the Jacobian.

\begin{figure}[ht]
  \includegraphics[width=.9\linewidth]{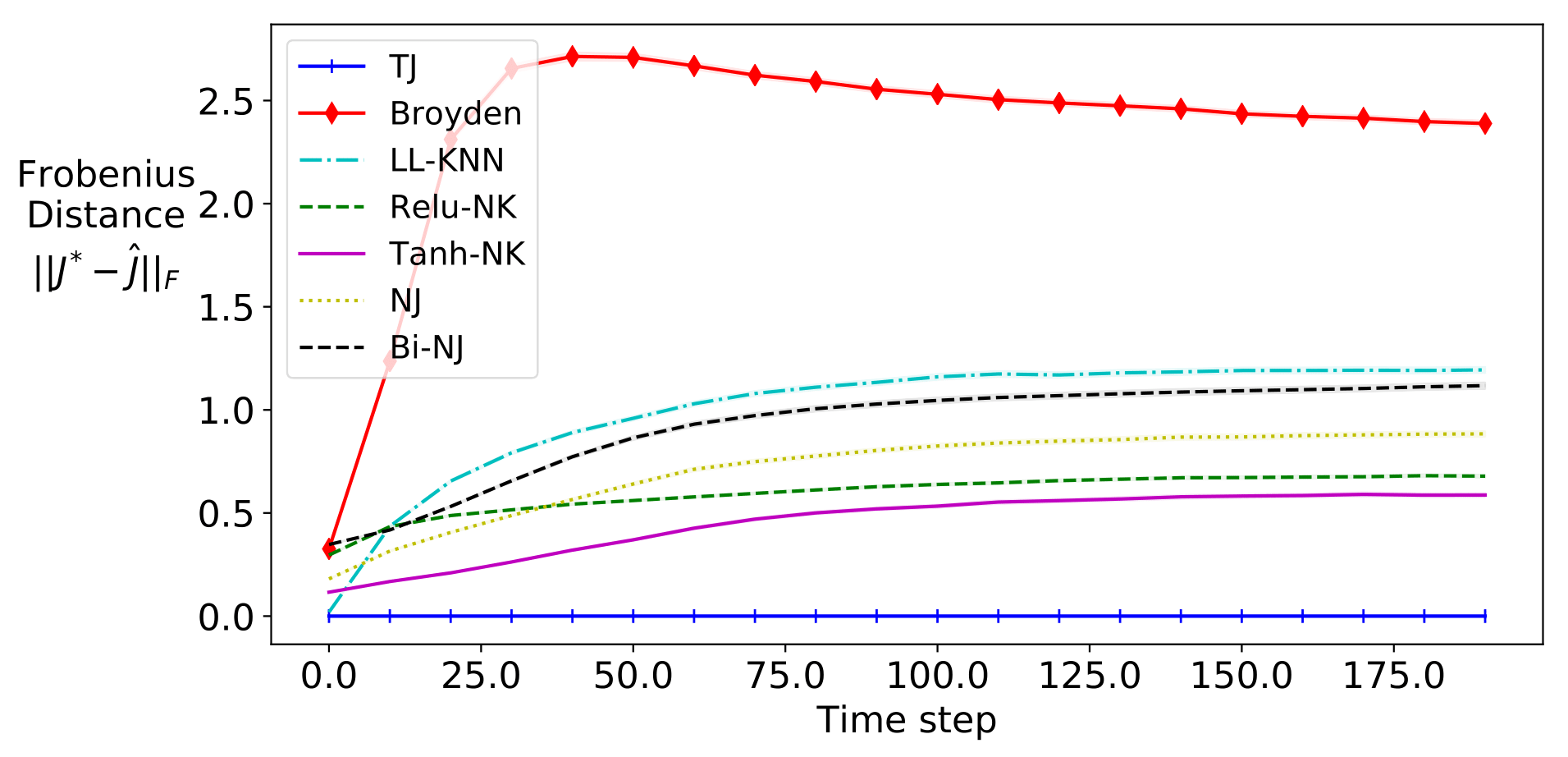}  
\caption{Frobenius distance of approximated Jacobians to the true Jacobian over trajectories. Plots are of mean Frobenius distances across trajectories, with standard-error of the mean.}
\label{fig:frobenius_errors_over_time}
\end{figure}

We report the Frobenius distance $||J^{*} - \hat{J}||_{F}$ in Figure~\ref{fig:frobenius_errors_over_time} average over time-steps of trajectories for our multi-point case. We again exclude the single-point results due to page constraints but observed similar trends in the multi-point scenario. All methods generally are less accurate over trajectories, which we accredit to the sparsity at more distal targets from the initial position because of our biased data collection process. More distal points from the initial position would have fewer training examples and likely worse approximations of the Jacobian in these regions. Our NK methods generally give closer approximations. Our Bidirectional Neural Jacobian is likely a less accurate approximation because the inverse Jacobian can have ambiguity in the relationship leading to different predictions from the true Jacobian. 

\begin{figure}[ht]
    \vspace{0.1in}
\begin{subfigure}{.45\textwidth}
  \centering
  \includegraphics[width=.9\linewidth]{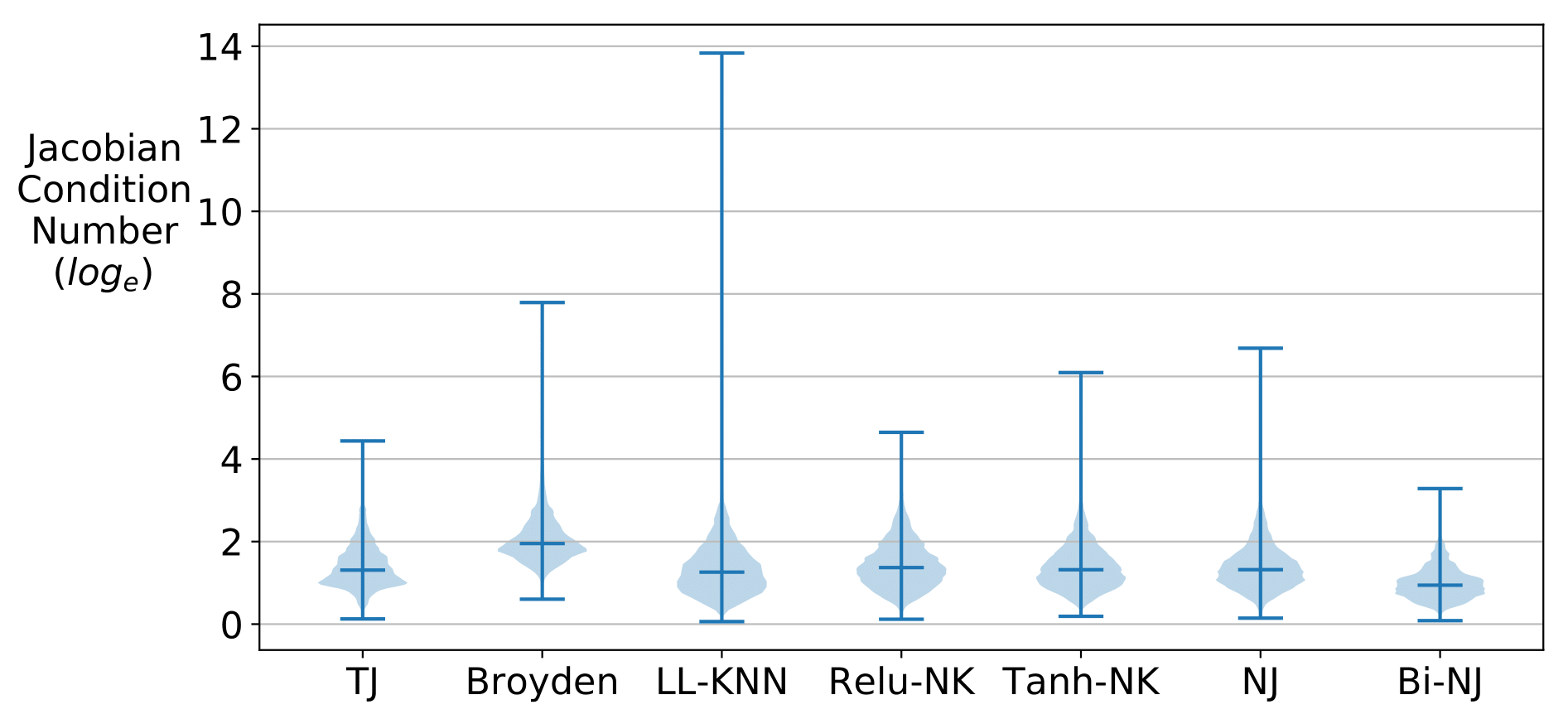}  
  \caption{Single Point Alignment}
  \label{fig:condition_single_pt}
\end{subfigure}
\begin{subfigure}{.45\textwidth}
  \centering
  \includegraphics[width=.9\linewidth]{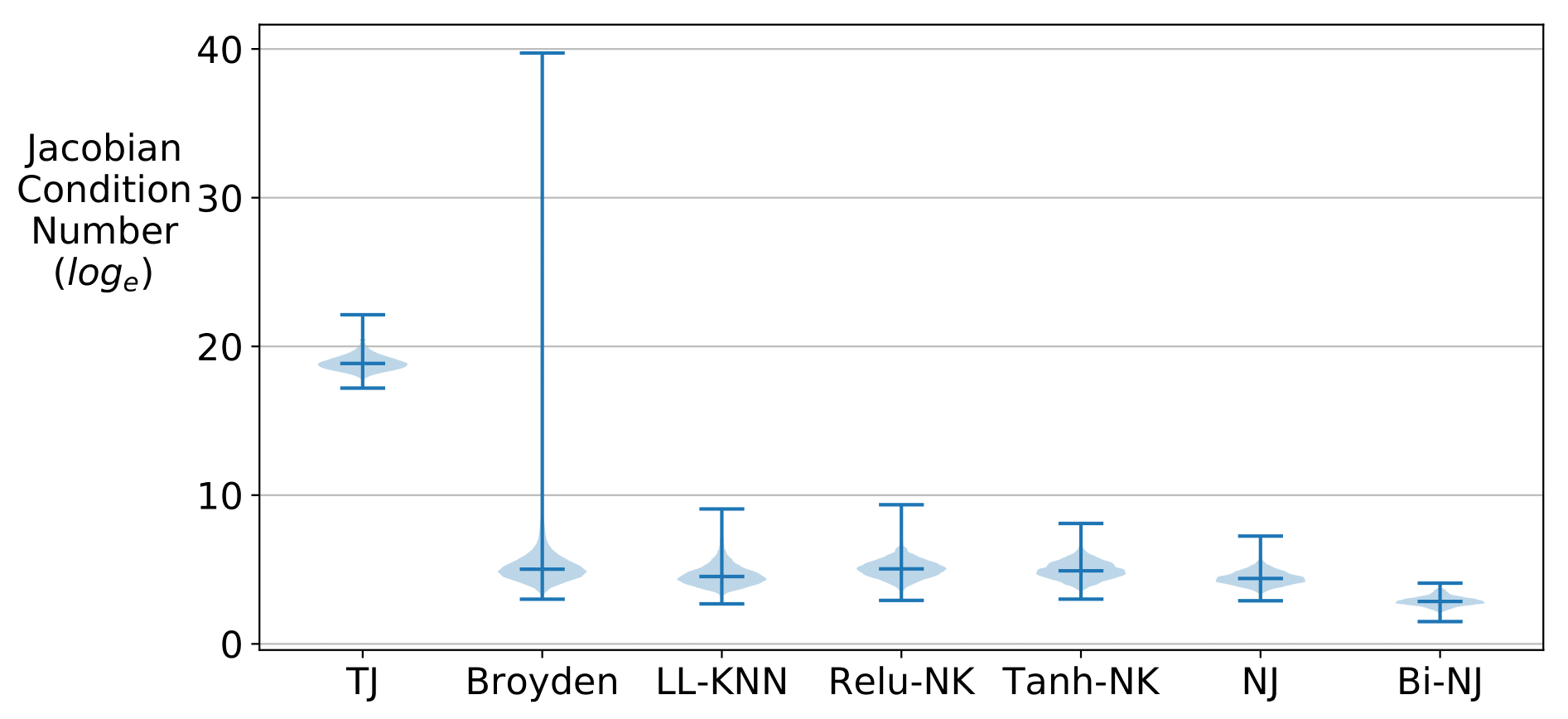}\caption{Multi-point alignment}
  \label{fig:condition_multi_pt}
\end{subfigure}
\caption{Violin plots of the natural log scale of the condition number of Jacobians from all simulation experiments. In }
\label{fig:condition_errors_over_time}
\end{figure}

We include our results on the condition number of our Jacobian matrices in Figure~\ref{fig:condition_errors_over_time}. Compared to the single-point scenario, the multi-point is inherently less well-conditioned and shown in our results. In the single point case, we generally find our neural methods better conditioned than the LL-KNN model, explaining the notable gap in performance. We also mention that we found the conditioning to be infinite in the LL-KNN case in several instances, which did not occur with our neural methods.  In the multi-point case, all learning methods are generally better conditioned than the true Jacobian, explaining why our neural methods perform better for closer targets. 

\begin{figure}[ht]
\hspace{2in}
    \centering
\includegraphics[width=0.95\linewidth]{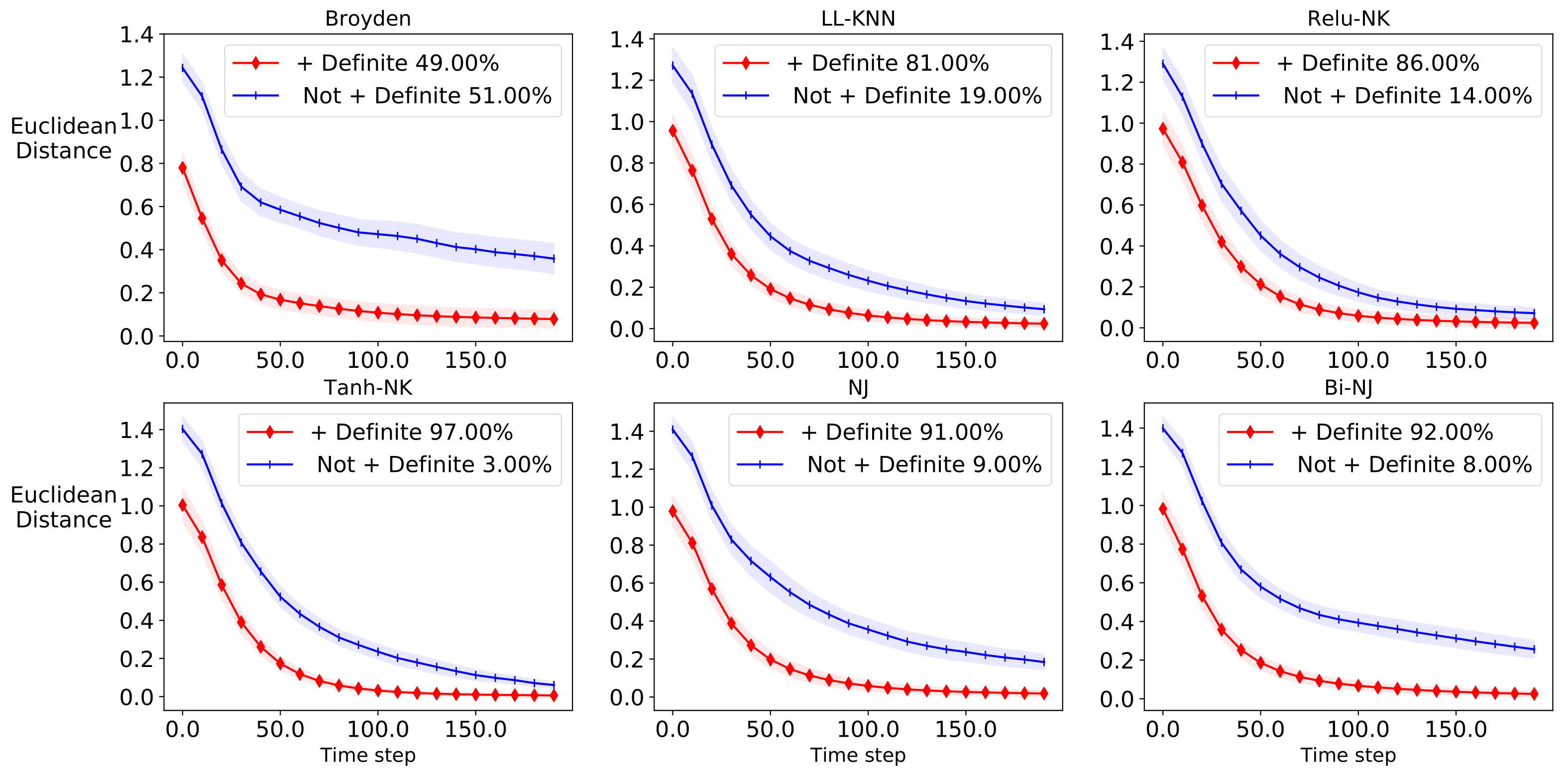}  
  \caption{Calculating the convergence criteria when calculated with true Jacobian matrix. When this criteria is not met the likelihood of failing to converge increases. Legend contains percentage of trajectories which were positive definite and not positive definite.}
    \label{fig:posdefcriteria_pt_to_pt}
\end{figure}

As one final analysis, in the single point environment, we check for positive definiteness between the true Jacobian $J^{*}$ multiplied by our approximations $\hat{J}$. Lyapunov theory says that if the Jacobian times its inverse is always positive definite ($J^{*} \hat{J}^{\dagger} \succ 0$) then our robot should converge locally to the target location  \citep{nematollahi2009generalizedcontrolawsIBVS}. We empirically explore this by checking for the positive definiteness of each Jacobian approximation in a trajectory. We then separate trajectories between always and not consistently positive definite results in Figure~\ref{fig:posdefcriteria_pt_to_pt}. Our results suggest that, on average, this criterion holds up in practice given the discrepancy between the successful and unsuccessful trajectories' final performance. Generally, our Bi-NJ and NK methods are the best to adhere to this criterion. We note that these results do not necessarily mean that a trajectory will not converge because the criterion is violated, but that there is no guarantee of convergence \citep{nematollahi2009generalizedcontrolawsIBVS}. Similarly, these conditions do not account for physical limitations, and some trajectories would converge under physical implausibilities, such as passing through the robot's joints.

\subsection{Real Robotic Experiments}

In this section, we test our neural methods on a physical Kinova Gen-3 light robotic manipulator. These experiments are primarily for a proof-of-concept that our methods can be applied to real robotic systems. We considered an image-based visual servoing task using a commercial webcam with a color hue tracker, and the other is a 7-DOF reaching task similar to our single point simulation. For each setting, we considered 100 randomly generated targets for evaluation. We intentionally excluded the true Jacobian as part of the evaluation because our focus is on data-driven Jacobian estimation for Cartesian control. We also do not evaluate the NK-Relu model because of the superior performance of our NK-Tanh model. We generate targets in a more constrained space of the Kinova's joint limits than our simulation experiments because of safety concerns. 

Table~\ref{tab:visualservosuccess} shows our visual servoing results. Our results show a similar trend where most methods outperform Broyden's method. However, unlike the simulation results, no single method is always better in performance. During data collection, the Kinova arm would sometimes go out-of-frame, introducing noisy training data where the relationship between the tracked point and joint angles was inaccurate. Our NK-Tanh method's performance drop is likely because it is trained to predict position from tracked points directly, and so tracking errors would directly affect these predictions. The performance drop suggests that further work is necessary to handle tracking errors with the NK method, such as training the model with a robust lost function. 

Table~\ref{tab:robosuccess} shows our 7 DOF robotic experiments. Our results seem to provide a similar ordering as our single point simulation results, but with a notable worse performance for the LL-KNN and Bi-NJ methods on the most distant targets, we mention that the LL-KNN method generally performs better with larger neighborhood sizes.

\begin{table}[h]%
\vspace{0.1in}
\centering
\setlength{\tabcolsep}{3 pt}
{ \small
\begin{tabular}{c|c|c|c|c|}
\cline{2-5}
\multicolumn{1}{l|}{}                & \multicolumn{4}{c|}{\textbf{Euclidean Distance}}                  \\ \hline
\multicolumn{1}{|c|}{\textbf{Model}} & 0.0 - 0.2      & 0.2 - 0.4      & 0.4 - 0.6      & 0.0  - 0.6     \\ \hline
\multicolumn{1}{|c|}{Broyden}        & 99.67          & 75.47          & 50.89          & 79.38          \\ \hline
\multicolumn{1}{|c|}{LL-KNN}         & \textbf{99.89} & 98.24          & 92.64          & 97.66          \\ \hline
\multicolumn{1}{|c|}{NJ}         & 99.43          & 97.59          & \textbf{96.73} & \textbf{98.10} \\ \hline
\multicolumn{1}{|c|}{Bi-NJ}      & 71.66          & 72.00          & 66.14          & 70.60          \\ \hline
\multicolumn{1}{|c|}{Tanh-NK}        & 99.70          & \textbf{98.52} & 87.73          & 96.64          \\ \hline
\end{tabular}
}
\caption{Percentage of successful trajectories reaching target on a 2 DOF visual servoing task. Percentages are averaged from a threshold of 0.001 - 0.1 Euclidean distance to target. }
\label{tab:visualservosuccess}
\end{table}

\begin{table}[h]
\centering
\setlength{\tabcolsep}{3 pt}
{
\begin{tabular}{c|c|c|c|c|c|}
\cline{2-6}
                                       & \multicolumn{5}{c|}{\textbf{Euclidean Distance (meters)}}                                                  \\ \hline
\multicolumn{1}{|c|}{\textbf{Model}}   & \textbf{0.0 - 0.25} & \textbf{0.25 - 0.5} & \textbf{0.5 - 0.75} & \textbf{0.75 - 1.3} & \textbf{0.0 - 1.3} \\ \hline
\multicolumn{1}{|c|}{\textbf{LL-KNN}}  & \textbf{100.0}      & 95.65               & 78.79               & 28.21               & 64.0               \\ \hline
\multicolumn{1}{|c|}{\textbf{NJ}}      & \textbf{100.0}      & 95.65               & 87.88               & 82.05               & 88.0               \\ \hline
\multicolumn{1}{|c|}{\textbf{Bi-NJ}}   & \textbf{100.0}      & 91.30               & 66.67               & 28.21               & 59.0               \\ \hline
\multicolumn{1}{|c|}{\textbf{Tanh-NK}} & \textbf{100.0}      & \textbf{100.00}     & \textbf{90.91}      & \textbf{84.62}      & \textbf{91.0}      \\ \hline
\end{tabular}
}
\caption{Percentage of successful trajectories reaching target on 7 DOF Kinova task. Percentages are averaged from a threshold of 0.001m - 0.1m Euclidean distance to target.}
\label{tab:robosuccess}
\end{table}



\section{Conclusion} 
\label{sec:conclusion}

In this work, we proposed two neural network approaches for learning the Jacobian matrix in robotic motion. Our experimental results showed that our neural methods performed better than our best data-driven baseline in several settings. In the more difficult multi-point case, we achieved a performance gain of 15.6\% for the Bidirectional Neural Jacobian and 20.3\% with our Neural Kinematics with Tanh over our best baseline results when considering all trajectories. We found that our approximations also gave a closer Jacobian approximation when compared to alternatives. On real robotics systems, our methods performed better than other algorithms in the robot's coordinate system. However, future work is necessary to understand how to best leverage the Neural Kinematics approach in visual servoing, where errors in the tracker negatively impact performance, as we observed in our experimental results. Another future direction is studying more specific cases, such as singularities, which is warranted for deploying our neural learning methods in applications.

\section*{Acknowledgement}
This work was sponsored by Canada CIFAR AI Chairs
Program, the Reinforcement Learning and Artificial
Intelligence (RLAI) Laboratory, and Alberta Machine
Intelligence Institute (Amii). We thank Amirmohammad Karimi for helping in setting up the robotic experiments.


\bibliography{bibliography}

\end{document}